\documentclass{article}

\usepackage[final]{neurips_2021}

\usepackage[utf8]{inputenc} 
\usepackage[T1]{fontenc}    
\usepackage{hyperref}       
\usepackage{url}            
\usepackage{booktabs}       
\usepackage{amsfonts}       
\usepackage{nicefrac}       
\usepackage{microtype}      
\usepackage{xcolor}         
\usepackage{multirow}
\usepackage{multicol}
\usepackage{amsmath}
\usepackage{algorithm}
\usepackage[noend]{algpseudocode}
\MakeRobust{\Call}
\usepackage{graphicx}

\title{SAE: Sequential Anchored Ensembles}

\author{%
  Arnaud Delaunoy\\
  University of Liège\\
  \texttt{a.delaunoy@uliege.be} \\
   \And
  Gilles Louppe\\
  University of Liège\\
  \texttt{g.louppe@uliege.be} \\
}

\DeclareMathOperator*{\argmax}{arg\,max}
\DeclareMathOperator*{\argmin}{arg\,min}

\DeclareMathOperator*{\x}{\boldsymbol{x}}
\DeclareMathOperator*{\y}{\boldsymbol{y}}
\DeclareMathOperator*{\D}{\boldsymbol{D}}
\DeclareMathOperator*{\te}{\boldsymbol{\theta}}

\begin{document}

\maketitle

\begin{abstract}
Computing the Bayesian posterior of a neural network is a challenging task due to the high-dimensionality of the parameter space. Anchored ensembles approximate the posterior by training an ensemble of neural networks on anchored losses designed for the optima to follow the Bayesian posterior. Training an ensemble, however, becomes computationally expensive as its number of members grows since the full training procedure is repeated for each member. In this note, we present Sequential Anchored Ensembles (SAE), a lightweight alternative to anchored ensembles. Instead of training each member of the ensemble from scratch, the members are trained sequentially on losses sampled with high auto-correlation, hence enabling fast convergence of the neural networks and efficient approximation of the Bayesian posterior. SAE outperform anchored ensembles, for a given computational budget, on some benchmarks while showing comparable performance on the others and achieved $2^\text{nd}$ and $3^\text{rd}$ place in the light and extended tracks of the NeurIPS 2021 Approximate Inference in Bayesian Deep Learning competition.
\end{abstract}

\section{Introduction}
Accurate uncertainty quantification has become of high importance for machine learning tasks where incorrect prediction could have severe consequences. Bayesian deep learning is a popular framework for capturing those uncertainties. It consists in estimating the Bayesian posterior
\begin{equation}
    p(\te |\D) = \frac{p(\D|\te) p(\te)}{p(\D)},
\end{equation}
where $\D$ is a dataset and $\te$ the parameters of the neural network. Predictions are then made through marginalization over the parameters
\begin{equation}
    p(\y|\x, \D) = \int p(\y | \x , \te) p(\te | \D) d\te,
\end{equation}
where $\y$ are the outputs and $\x$ are the inputs. In practice, inferring the Bayesian posterior is challenging due to the high-dimensionality of the parameter space; scalable techniques to approximate this posterior are hence needed. 

One of such techniques is called anchored ensembles \citep{pearce2020uncertainty}. It builds an ensemble of neural networks with a randomized objective function designed for the optima to approximately follow the Bayesian posterior $p(\te \vert \D)$. Under the assumption of a normal prior $p(\te) = \mathcal{N}(\mu_{\text{prior}}, \Sigma_{\text{prior}})$ and likelihood $p(\D \vert \te)$, the optima $\te^*$ obtained by minimizing the anchored loss $-(\log p(\D \vert \te) + \log p_{\text{anc}}(\te))$, where $p_{\text{anc}} = \mathcal{N}(\te_{\text{anc}}, \Sigma_{\text{prior}})$, approximately follow the Bayesian posterior $p(\te|\D)$ if $\te_{\text{anc}} \sim p(\te)$.

\begin{center}
\begin{minipage}[t]{0.85\textwidth}
\begin{algorithm}[H]
\caption{Anchored Ensembling (AE)}\label{alg:ae}
\begin{algorithmic}
\For{$i$ in $1, .., N$}
\State $\te_{\text{anc}, i} \sim p(\te)$ \Comment{Sample anchor}
\State $\te_{\text{init}, i} \gets \Call{init}{ }$ \Comment{Initialize NN}
\State $\te^*_i \gets  \Call{train}{\te_{\text{anc}, i}, \te_{\text{init}, i}}$ \Comment{$\te^*_i \gets \argmin_{\te} -(\log p(\D \vert \te) + \log p_{\text{anc}}(\te))$}
\EndFor
\end{algorithmic}
\end{algorithm}
\end{minipage}
\end{center}

\section{Sequential anchored ensembles}

\begin{figure}[H]
    \centering
    \includegraphics[width=\textwidth]{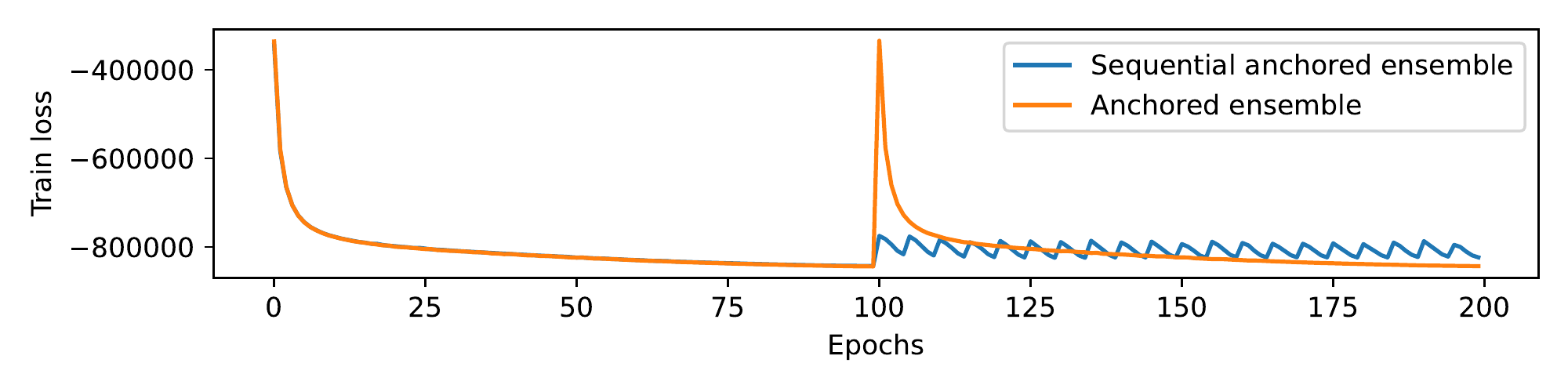}
    \caption{While anchored ensembles start the training from scratch at a high loss, sequential anchored ensembles start from the previously obtained solution and hence at a lower loss allowing to build an ensemble of 21 members in the time anchored ensembles built an ensemble of 2 members. Each peak corresponds to the start of a new training procedure, the fast convergence of sequential anchored ensembles allows to train much more ensemble members than anchored ensembles in the same time.}
    \label{fig:losses}
\end{figure}

In this work, we propose Sequential Anchored Ensembles (SAE) which stem from the observation that if $\te_{\text{anc}, i-1}$ and $\te_{\text{anc}, i}$ are close so should be $\te^*_{i-1}$ and $\te^*_{i}$. Therefore, training all the members of the ensemble from scratch is inefficient. We exploit this to reduce the computational cost of the training procedure by training the elements of the ensemble sequentially starting from the previous solution $\te^*_{i-1}$. This is illustrated in Figure \ref{fig:losses} and summarized in Algorithm \ref{alg:sae}. 

Let us consider the construction of an ensemble composed of $N$ members $\te^*_1, .., \te^*_N$, with $\te^*_i$ constructed based on the anchor $\te_{\text{anc}, i}$. A first anchor is sampled from the prior and the corresponding neural network is trained in a classical way. The next anchors are sampled such that two consecutive anchors are close to each other and the corresponding neural networks are trained starting from the previous optimum. As consecutive anchors are close, so are the solutions, and the training procedure is then expected to converge much faster. To sample consecutive anchors that are close to each other but eventually span the prior, we rely on an MCMC procedure. As the parameters are assumed independent under the prior, we run a separate chain for each parameter of the neural network so that some transitions are accepted by the Metropolis-Hastings algorithm at each step. The parameters of the anchors being independently normally distributed, the distribution is easy to navigate as opposed to an MCMC procedure directly performed on the posterior. To benefit from parallelization and decrease ensemble correlation, the algorithm can be run multiple times with different initializations.

\begin{center}
\begin{minipage}[t]{0.85\textwidth}
\begin{algorithm}[H]
\caption{Sequential Anchored Ensembling (SAE)}\label{alg:sae}
\begin{algorithmic}
\State $\te_{\text{anc}, 1} \sim p(\te)$ \Comment{Sample first anchor}
\State $\te_{\text{init}, 1} \gets \text{init}()$ \Comment{Initialize NN}
\State $\boldsymbol{d}_1 \gets \Call{RandChoice}{\Call{size}{\te_{\text{anc}, 1}}, \{-1, 1\}}$ \Comment{Direction}
\State $\te^*_1 \gets \Call{train}{\te_{\text{anc}, 1}, \te_{\text{init}, 1}}$ \Comment{Long training}
\For{$i$ in $2, .., N$} \Comment{$i^{\text{th}}$ step of the SAE algorithm}
\State $\te_{\text{anc}, i}, \boldsymbol{d}_i \gets$ \Call{MH\_update}{$(\te_{\text{anc}, i-1}, \boldsymbol{d}_{i-1}$} \Comment{Alg. \ref{alg:guided_walk}}
\State $\te_{\text{init}, i} \gets \te^*_{i-1}$ \Comment{Start from previous optimum}
\State $\te^*_i \gets \Call{train}{\te_{\text{anc}, i}, \te_{\text{init}, i}}$ \Comment{Short training}
\EndFor
\end{algorithmic}
\end{algorithm}
\end{minipage}
\end{center}

\begin{center}
\begin{minipage}[t]{0.85\textwidth}
\begin{algorithm}[H]
\caption{Guided walk Metropolis-Hastings}\label{alg:guided_walk}
\begin{algorithmic}
\Function{MH\_update}{$\te_{\text{anc}}$, $\boldsymbol{d}$}
\State $\te'_{\text{anc}} \gets \Call{empty}{\Call{size}{\te_{\text{anc}}}}$, $\boldsymbol{d}' \gets \Call{empty}{\Call{size}{\boldsymbol{d}}}$
\For{$(\theta_{\text{anc}, j}, d_j)$ in $(\te_{\text{anc}}, \boldsymbol{d})$} \Comment{$j^{\text{th}}$ anchor's parameter}
\State $y \gets \theta_{\text{anc}, j} + d_j |z|, \quad z \sim \mathcal{N}(0, \sigma_{\text{step}})$ \Comment{Transition}
\State $\alpha \gets \min\left( \frac{p(y)}{p(\theta_{\text{anc}, j})}, 1 \right)$ \Comment{Acceptance probability}
\State $u \sim \mathcal{U}(0, 1)$
\If{$u < \alpha$} \Comment{Accept transition}
    \State $\theta_{\text{anc}, j}' \gets y, \quad d_j' \gets d_j$ \Comment{Keep same direction}
\Else \Comment{Reject transition}
    \State $\theta_{\text{anc}, j}' \gets \theta_{\text{anc}, j}, \quad d_j' \gets -d_j$ \Comment{Invert direction}
\EndIf
\EndFor
\State\Return $\te'_{\text{anc}}, \boldsymbol{d}'$
\EndFunction
\end{algorithmic}
\end{algorithm}
\end{minipage}
\end{center}

The algorithm can be performed with any MCMC procedure, however, to be efficient, the MCMC procedure should eventually span the whole prior space so that the optima span the whole posterior while making small transitions for training to remain fast. We have found that a guided walk Metropolis-Hastings procedure with Gaussian transitions \citep{gustafson1998guided} performs well. This procedure is illustrated in Algorithm \ref{alg:guided_walk}. The difference between a classical Metropolis-Hastings algorithm and its guided walk version is that the latter always performs transitions in the same direction until a rejection occurs. On rejection, the transition direction is inverted until the next rejection. As we know the prior to be normally distributed, rejection cannot occur until the point of maximal density is passed. The anchors will then consistently evolve in the same direction hence allowing to efficiently span the prior with short steps.

\section{Experiments}
The code used to conduct the experiments is available at \url{https://github.com/ADelau/SAE-Sequential-Anchored-Ensembles}. To assess the efficiency of SAE, we compare how close the approximated predictive density $1/N \sum_{i=1}^N p(\y|\x, \te^*_i)$ is to the true density $\int p(\y | \x, \te) p(\te | \D) d\te$ both for anchored ensembles and sequential anchored ensembles. For a given computational budget, the sequential anchored ensembles will be composed of more members than the anchored ensembles but each member will be trained for a shorter time and built sequentially. The true posterior used for comparison has been computed by a Hamiltonian Monte-Carlo procedure \citep{izmailov2021bayesian} in the context of the NeurIPS 2021 Approximate Inference in Bayesian Deep Learning competition \citep{wilson2021evaluating}.

\renewcommand{\arraystretch}{1.4}
\begin{table}[h!]
    \centering
    \resizebox{\columnwidth}{!}{%
    \begin{tabular}{|c|c|c|c|c|c|c|c|c|c|c|}
    \cline{3-11}
    \multicolumn{2}{c|}{} & \multicolumn{2}{|c|}{Cifar10 Resnet-20} & \multicolumn{2}{c|}{Cifar10(-C) Alexnet} & \multicolumn{2}{c|}{IMDB} & \multicolumn{2}{c|}{DermaMNIST} & UCI-Gap \\
    \cline{3-11}
    \multicolumn{2}{c|}{} & Ag. & TV & Ag. & TV & Ag. & TV & Ag. & TV & $W_2$\\
    \hline
    $1000$ & AE & $0.849$ & $0.201$ & $0.726$ & $0.262$ & $\mathbf{0.892}$ & $\mathbf{0.109}$ & $0.877$ & $0.104$ & $\mathbf{0.148}$\\
    \cline{2-11}
    epochs & SAE & $\mathbf{0.856}$ & $\mathbf{0.176}$ & $\mathbf{0.772}$ & $\mathbf{0.212}$ & $0.887$ & $0.110$ & $\mathbf{0.880}$ & $\mathbf{0.098}$ & $0.159$\\
    \hline
    $10,000$ & AE & $0.862$ & $0.199$ & $0.746$ & $0.236$ & $\mathbf{0.926}$ & $\mathbf{0.086}$ & $\mathbf{0.897}$ & $0.089$ & $\mathbf{0.137}$ \\
    \cline{2-11}
    epochs & SAE & $\mathbf{0.903}$ & $\mathbf{0.133}$ & $\mathbf{0.787}$ & $\mathbf{0.200}$ & $0.916$ & $0.099$ & $0.893$ & $\mathbf{0.086}$ & $0.143$ \\
    \hline
    \end{tabular}
    }
    \caption{Comparison of the performance of anchored ensembles and sequential anchored ensembles. While the computational budget is split uniformly between ensemble members for anchored ensembles, sequential anchored ensembles perform many short training procedures allowing to construct more ensemble members for the same computational budget. The reported values are the median performance over at least 20 runs. Cifar10(-C) corresponds to a dataset composed of $20\%$ of uncorrupted samples and $80\%$ of samples with a corruption level of 4.}
    \label{tab:results}
\end{table}
\renewcommand{\arraystretch}{1.}

Results for computational budgets of $1000$ and $10,000$ epochs are shown in Table \ref{tab:results}. Additional results, experimental details and metrics definitions can be found in Appendix \ref{add_exp}. We observe that for a given computational budget, SAE performs better than AE on Cifar10 and Cifar10-C, slightly worse on UCI-Gap and shows similar performance on the other datasets. It shows that the sequential procedure is able to navigate the posterior and sometimes approximates the posterior density more efficiently than traditional anchored ensembles. Interestingly, we observe that when both methods yield a similar agreement, SAE usually yields a lower total variation. SAE hence tends to provide better-calibrated posteriors even when classification performance is similar. The reason why it performs better on some datasets than others is still unclear. However, we hypothesize that some posteriors are easier to navigate leading to high performance using SAE while others are harder and benefit more from different initializations of the weights.

\section{Related work}
Building an ensemble in a sequential manner has already been proven successful in the context of non-anchored deep ensembles. Snapshot ensembles \citep{huang2017snapshot} and Fast Geometric Ensembling \citep{garipov2018loss} build a deep ensemble by recording weights during the training procedure. They make use of a cyclical learning rate schedule to increase diversity. Stochastic Weights Averaging \citep{izmailov2018averaging} directly averages the weights in place of the predictions leading to lower memory requirements and prediction complexity. Sequential anchored ensembling follows similar ideas but uses anchors to create diversity in the ensemble following the Bayesian posterior. Improvements of those methods include low-precision training \citep{yang2019swalp} and Simplicial Pointwise Random Optimization \citep{benton2021loss} which build simplexes of low loss. Such ideas could possibly be adapted for the sequential anchored ensembles setting, improving its performance. 

\begin{ack}
The authors would like to thank Tim Pearce for his insightful comments and feedback. Arnaud Delaunoy would like to thank the National Fund for Scientific Research (F.R.S.-FNRS) for his scholarship. Gilles Louppe is recipient of the ULiège - NRB Chair on Big Data and is thankful for the support of the NRB.
\end{ack}

\bibliographystyle{plainnat}
\bibliography{main}


\appendix

\section{Additional experiments}\label{add_exp}
In this section, we provide results on smaller budgets and all the experimental details, results are shown in Table \ref{tab:add_results}. Details about the architectures can be found in Table \ref{tab:architectures}. The allocation of the computational budget for the UCI-Gap dataset can be found in Table \ref{tab:budget_gap} while the others can be found in Table \ref{tab:budget}. The metrics used to compare the approximate posterior to the true one are, for classification, the agreement
\begin{equation}
    \text{agreement}(p, \hat{p}) = \frac{1}{n} \sum_{i=1}^{N} I[\argmax_j \hat{p}(y_j|x_i) = \argmax_j p(y_j|x_i)]
\end{equation}
and the total variation
\begin{equation}
    \text{TV}(p, \hat{p}) = \frac{1}{n}\sum_{i=1}^n \frac{1}{2} \sum_j \left| p(y_j|x_i) - \hat{p}(y_j|x_i) \right|,
\end{equation}
where $p$ is the true predictive density and $\hat{p}$ is the approximate one. For regression we use a point-wise Wasserstein-2 distance based on samples from the predictive distribution
\begin{equation}
    W_2(p, \hat{p}) = \inf_I \sqrt{\sum_{i \in I, j}|p_i - \hat{p}_j|^2}.
\end{equation}

\renewcommand{\arraystretch}{1.4}
\begin{table}[h!]
    \centering
    \resizebox{\columnwidth}{!}{%
    \begin{tabular}{|c|c|c|c|c|c|}
        \hline
        \multirow{2}*{Agreement} & budget & \multirow{2}*{$200$} & \multirow{2}*{$500$} & \multirow{2}*{$1000$} & \multirow{2}*{$10,000$}\\
         & (epochs) & & & & \\
        \hline
        \hline
        \multirow{2}*{Cifar10 Resnet-20} & AE & $\mathbf{0.804}^{+0.010}_{-0.020}$ & $\mathbf{0.833}^{+0.009}_{-0.015}$ & $0.849^{+0.006}_{-0.011}$ & $0.862^{+0.005}_{-0.005}$ \\
        \cline{2-6}
        & SAE & $0.782^{+0.012}_{-0.025}$ & $0.830^{+0.010}_{-0.017}$ & $\mathbf{0.856}^{+0.006}_{-0.010}$ & $\mathbf{0.903}^{+0.004}_{-0.002}$ \\
        \hline
        \hline
        \multirow{2}*{Cifar10-C Alexnet} & AE & $0.660^{+0.007}_{-0.010}$ & $0.706^{+0.005}_{-0.007}$ & $0.726^{+0.004}_{-0.003}$ & $0.746^{+0.003}_{-0.007}$ \\
        \cline{2-6}
        & SAE & $\mathbf{0.738}^{+0.006}_{-0.014}$ & $\mathbf{0.762}^{+0.003}_{-0.005}$ & $\mathbf{0.772}^{+0.005}_{-0.004}$ & $\mathbf{0.787}^{+0.002}_{-0.003}$ \\
        \hline
        \hline
        \multirow{2}*{IMDB} & AE & $0.829^{+0.008}_{-0.009}$ & $0.865^{+0.007}_{-0.008}$ & $\mathbf{0.892}^{+0.004}_{-0.003}$ & $\mathbf{0.926}^{+0.001}_{-0.002}$ \\
        \cline{2-6}
        & SAE & $\mathbf{0.831}^{+0.003}_{-0.005}$ & $\mathbf{0.868}^{+0.004}_{-0.006}$ & $0.887^{+0.004}_{-0.007}$ & $0.916^{+0.002}_{-0.001}$ \\
        \hline
        \hline
        \multirow{2}*{DermaMNIST} & AE & $0.826^{+0.014}_{-0.021}$ & $0.865^{+0.011}_{-0.008}$ & $0.877^{+0.007}_{-0.006}$ & $\mathbf{0.897}^{+0.005}_{-0.009}$ \\
        \cline{2-6}
        & SAE & $\mathbf{0.836}^{+0.009}_{-0.027}$ & $\mathbf{0.869}^{+0.011}_{-0.008}$ & $\mathbf{0.880}^{+0.006}_{-0.012}$ & $0.893^{+0.008}_{-0.014}$ \\
        \hline
        
        \noalign{\vskip 4mm} 
        \hline
        Total variation & budget & $200$ & $500$ & $1000$ & $10,000$\\
        \hline
        \hline
        \multirow{2}*{Cifar10 Resnet-20} & AE & $0.225^{+0.017}_{-0.014}$ & $0.211^{+0.015}_{-0.007}$ & $0.201^{+0.012}_{-0.011}$ & $0.199^{+0.003}_{-0.002}$ \\
        \cline{2-6}
        & SAE & $\mathbf{0.223}^{+0.022}_{-0.012}$ & $\mathbf{0.194}^{+0.020}_{-0.012}$ & $\mathbf{0.176}^{+0.011}_{-0.007}$ & $\mathbf{0.133}^{+0.002}_{-0.003}$ \\
        \hline
        \hline
        \multirow{2}*{Cifar10-C Alexnet} & AE & $0.335^{+0.005}_{-0.008}$ & $0.283^{+0.005}_{-0.007}$ & $0.262^{+0.006}_{-0.003}$ & $0.236^{+0.006}_{-0.002}$ \\
        \cline{2-6}
        & SAE & $\mathbf{0.245}^{+0.014}_{-0.006}$ & $\mathbf{0.221}^{+0.004}_{-0.003}$ & $\mathbf{0.212}^{+0.004}_{-0.004}$ & $\mathbf{0.200}^{+0.004}_{-0.003}$ \\
        \hline
        \hline
        \multirow{2}*{IMDB} & AE & $0.180^{+0.007}_{-0.006}$ & $0.131^{+0.005}_{-0.003}$ & $\mathbf{0.109}^{+0.003}_{-0.005}$ & $\mathbf{0.086}^{+0.001}_{-0.001}$ \\
        \cline{2-6}
        & SAE & $\mathbf{0.160}^{+0.003}_{-0.003}$ & $\mathbf{0.124}^{+0.004}_{-0.003}$ & $0.110^{+0.004}_{-0.003}$ & $0.099^{+0.001}_{-0.002}$ \\
        \hline
        \hline
        \multirow{2}*{DermaMNIST} & AE & $0.153^{+0.012}_{-0.014}$ & $0.117^{+0.006}_{-0.012}$ & $0.104^{+0.007}_{-0.006}$ & $0.089^{+0.005}_{-0.009}$ \\
        \cline{2-6}
        & SAE & $\mathbf{0.143}^{+0.023}_{-0.014}$ & $\mathbf{0.111}^{+0.006}_{-0.010}$ & $\mathbf{0.098}^{+0.011}_{-0.007}$ & $\mathbf{0.086}^{+0.004}_{-0.001}$ \\
        \hline
        
        \noalign{\vskip 4mm} 
        \hline
        $W_2$ & budget & $200$ & $500$ & $1000$ & $10,000$\\
        \hline
        \hline
        \multirow{2}*{UCI-Gap} & AE & $\mathbf{0.174}^{+0.357}_{-0.054}$ & $\mathbf{0.152}^{+0.083}_{-0.042}$ & $\mathbf{0.148}^{+0.142}_{-0.032}$ & $\mathbf{0.137}^{+0.039}_{-0.021}$ \\
        \cline{2-6}
        & SAE & $0.196^{+0.481}_{-0.072}$ & $0.169^{+0.149}_{-0.059}$ & $0.159^{+0.162}_{-0.068}$ & $0.143^{+0.096}_{-0.041}$ \\
        \hline
    \end{tabular}
    }
    \caption{Comparison of the performance of anchored ensembles and sequential anchored ensembles for various computational budgets. While the computational budget is split uniformly between ensemble members for anchored ensembles, sequential anchored ensembles perform many short training procedures allowing to construct more ensemble members for the same computational budget. The reported values are the median, minimal and maximal performance over 100 runs for the UCI-Gap dataset and over 20 runs for the other datasets.}
    \label{tab:add_results}
\end{table}

\begin{table}[]
    \centering
    \begin{tabular}{|c|c|}
        \hline
        Cifar10 Resnet-20 & 20-layers CNN \\
        \hline
        Cifar10 Alexnet & 5-layers CNN \\
        \hline
        IMDB & 1 Conv layer + LSTM cell \\
        \hline
        DermaMNIST & 2-layers CNN \\
        \hline
        UCI-Gap & 2-layers MLP \\
        \hline
    \end{tabular}
    \caption{Architectures used with each dataset.}
    \label{tab:architectures}
\end{table}

\begin{table}[]
    \centering
    \resizebox{\columnwidth}{!}{%
    \begin{tabular}{|c|c|c|c|c|}
        \hline
        Budget (epochs) & $200$ & $500$ & $1000$ & $10,000$ \\
        \hline
        \multirow{2}{*}{AE} & 2 members of $100$ epochs & $5 \times 100$& $10 \times 100$ & $100 \times 100$ \\
        & $2$ members & $5$ members & $10$ members & $100$ members \\
        \hline
        \multirow{3}{*}{SAE} & $1$ chain of $100$ first epochs & \multirow{2}{*}{$2(100 + 75 \times 2)$} & \multirow{2}{*}{$3(100 + 116 \times 2)$} & \multirow{2}{*}{$10(100 + 450 \times 2)$} \\
         & $50$ sequential trainings of $2$ epochs & & & \\
         & $51$ members & $152$ members & $351$ members & $4510$ members \\
        \hline
    \end{tabular}
    }
    \caption{Allocation of the computational budget for the Cifar10, Cifar10(-C), IMDB and DermaMNIST datasets.}
    \label{tab:budget}
\end{table}

\begin{table}[]
    \centering
    \resizebox{\columnwidth}{!}{%
    \begin{tabular}{|c|c|c|c|c|}
        \hline
        Budget (epochs) & $200$ & $500$ & $1000$ & $10,000$ \\
        \hline
        \multirow{2}{*}{AE} & 2 members of $100$ epochs & $5 \times 100$& $10 \times 100$ & $100 \times 100$ \\
        & $2$ members & $5$ members & $10$ members & $100$ members \\
        \hline
        \multirow{3}{*}{SAE} & $1$ chain of $100$ first epochs & \multirow{2}{*}{$2(100 + 15 \times 10)$} & \multirow{2}{*}{$3(100 + 23 \times 10)$} & \multirow{2}{*}{$10(100 + 90 \times 10)$} \\
         & $10$ sequential trainings of $10$ epochs & & & \\
         & $11$ members & $32$ members & $72$ members & $910$ members \\
        \hline
    \end{tabular}
    }
    \caption{Allocation of the computational budget for the UCI-Gap dataset.}
    \label{tab:budget_gap}
\end{table}

\renewcommand{\arraystretch}{1.}

\end{document}